\def\year{2021}\relax
\documentclass[letterpaper]{article} 
\usepackage{aaai21}  
\usepackage{times}  
\usepackage{helvet} 
\usepackage{courier}  
\usepackage[hyphens]{url}  
\usepackage{graphicx} 
\usepackage{natbib}  
\usepackage{caption} 
\usepackage{amsmath,amsfonts,bm}

\def\eqref#1{equation~\ref{#1}}

\def\1{\bm{1}}

\DeclareMathAlphabet{\mathsfit}{\encodingdefault}{\sfdefault}{m}{sl}
\SetMathAlphabet{\mathsfit}{bold}{\encodingdefault}{\sfdefault}{bx}{n}

\def\gB{{\mathcal{B}}}

\def\gD{{\mathcal{D}}}

\def\gI{{\mathcal{I}}}

\def\sR{{\mathbb{R}}}

\def\sZ{{\mathbb{Z}}}

\usepackage{amsmath}
\usepackage{multirow}
\usepackage{booktabs}

\title{Learning Pseudo-Backdoors for Mixed Integer Programs}

\author {
    Aaron Ferber, \textsuperscript{\rm 1}
    Jialin Song, \textsuperscript{\rm 2}
    Bistra Dilkina,\textsuperscript{\rm 1}
    Yisong Yue \textsuperscript{\rm 2}\\
}
\affiliations {
    \textsuperscript{\rm 1} University of Southern California \\
    \textsuperscript{\rm 2} California Institute of Technology\\
    aferber@usc.edu, jssong@caltech.edu, bdilkina@usc.edu, yyue@caltech.edu
}

\begin{document}

\maketitle

\begin{abstract}
We propose a machine learning approach for quickly solving Mixed Integer Programs (MIP) by learning to prioritize a set of decision variables, which we call pseudo-backdoors, for branching that results in faster solution times. Learning-based approaches have seen success in the area of solving combinatorial optimization problems by being able to flexibly leverage common structures in a given distribution of problems. Our approach takes inspiration from the concept of strong backdoors, which corresponds to a small set of variables such that only branching on these variables yields an optimal integral solution and a proof of optimality.  Our notion of pseudo-backdoors corresponds to a small set of variables such that only branching on them leads to faster solve time (which can be solver dependent).  A key advantage of pseudo-backdoors over strong backdoors is that they are much amenable to data-driven identification or prediction. Our proposed method learns to estimate the solver performance of a proposed pseudo-backdoor, using a labeled dataset collected on a set of training MIP instances. This model can then be used to identify high-quality pseudo-backdoors on new MIP instances from the same distribution. We evaluate our method on the generalized independent set problems and find that our approach can efficiently identify high-quality pseudo-backdoors. In addition, we compare our learned approach against Gurobi, a state-of-the-art MIP solver, demonstrating that our method can be used to improve solver performance.
\end{abstract}

\section{Introduction}
\label{sec:intro}

Mixed integer programs (MIPs) are widely used mathematical models for combinatorial optimization problems \citep{conforti2014integer} that are generally solved to optimality with a tree search algorithm called branch-and-bound \citep{land2010automatic}.
Backdoors, initially introduced for SAT \citep{williams2003backdoors, dilkina2009backdoorsat} and then generalized to MIPs \citep{ 
dilkina2009backdoorsinitial}, are defined as subsets of integer variables such that only branching on them yields an optimal integral solution and a certificate of optimality.
While their contribution to SAT has theoretical and practical limitations \cite{ jarvisalo2007bdlimitations, jarvisalo2008bdeffect, semenov2018adversarialcryptographic}, solve time speedup in MIPs has been observed by prioritizing ``backdoors'' in branching \citep{fischetti2011backdoor}. We consider a subset of integer variables a \textit{pseudo-backdoor} if prioritizing those variables leads to faster MIP solve time. 

We introduce a data-driven approach to predicting pseudo-backdoors for distributions of MIPs with a \textit{scoring model} to identify high-quality pseudo-backdoors among a sample of candidates, and a subsequent \textit{classification model} to decide whether to use a candidate pseudo-backdoor by setting branching priority accordingly or just use the default solver. 
We conduct empirical evaluations on the Generalized Independent Set Problem (GISP) \citep{colombi2017generalized} and show that our models achieve faster solve times than Gurobi.

\begin{table*}[ht]
\centering
\begin{tabular}{rrcccccc}
\toprule
dataset     & solver     & \multicolumn{1}{l}{mean} & \multicolumn{1}{l}{stdev} & \multicolumn{1}{l}{25 pct} & \multicolumn{1}{l}{median} & \multicolumn{1}{l}{75 pct} & win / tie / loss vs grb \\
\midrule

gisp easy & grb        & 611                      & \textbf{182}              & 488                        & 580                        & 681                        & 0 / 100 / 0             \\
gisp easy & scorer       & 960                      & 755                       & 515                        & 649                        & 915                        & 41 / 0 / 59             \\
gisp easy & scorer + cls & \textbf{601}             & 247                       & \textbf{481}               & \textbf{568}               & \textbf{663}               & 24 / 70 / 6             \\
\midrule
gisp hard   & grb        & 2533                     & 939                       & 1840                       & 2521                       & 2976                       & 0 / 100 / 0             \\
gisp hard   & scorer       & 2373                     & \textbf{855}              & 1721                       & 2262                       & 2926                       & 47 / 0 / 53             \\
gisp hard   & scorer + cls & \textbf{2326}            & \textbf{855}              & \textbf{1654}              & \textbf{2215}              & \textbf{2866}              & 47 / 27 / 26           \\
\bottomrule
\end{tabular}
\caption{Runtime comparison in seconds of standard gurobi (grb), the score model (scorer), and the score model with subsequent classification (scorer+cls) on the test set for 2 hardness settings of gisp. }
\label{tab:results}

\end{table*}
\section{Learning Pseudo-Backdoors for MIP}

Our goal in finding pseudo-backdoors is to quickly solve MIPs. In a MIP we are asked to find real-valued settings for $n$ decision variables $x \in \sR^n$, which maximize a linear objective function $c^Tx$, subject to $m$ linear constraints $Ax\leq b$, and with a subset $\gI \subseteq [n]$ of the decision variables required to be integral $x_i \in \sZ \,\forall i \in \gI$. The problem can be written as $\max_x \{c^Tx : Ax \leq b,  x_i  \in \sZ \, \forall i \in \gI\}$.

Given a MIP, specified by $P=(c, A, b, \gI)$, we want to find a pseudo-backdoor subset $\gB \subseteq \gI$, of the integral decision variables such that prioritizing branching on these decision variables yields fast solve times. We consider a distributional setting of MIP solving where we train a model on several MIP instances from a distribution and deploy it on unseen instances from the same distribution. 

Our approach uses two learned models: one that scores subsets of variables according to performance as pseudo-backdoors, and another that classifies whether to use a given pseudo-backdoor over a standard solver. The score model $S(P, \gB; \theta_S)$ is parametrized by neural network parameters $\theta_S$ which takes as input the MIP specification $P$, and a candidate subset $\gB$, then predicts a score that characterizes if $\gB$ is a good pseudo-backdoor. The classifier $C(P, \gB; \theta_C)$ is parametrized by neural network parameters $\theta_C$ and predicts whether  prioritizing $\gB$ in branching would produce a smaller runtime than a standard solver. 
Our models use Graph Attention Network \cite{velickovic2018graph} to perform message passing followed by global attention pooling \cite{li2016gated} to represent the graph as a single feature vector, which is then fed into a fully connected network to produce scalar predictions.
The MIP is represented as a bipartite graph as in \cite{gasse2019exact} for permutation invariance and parameter sharing. Each MIP has two sets of nodes, one with with nodes representing variables and another representing constraints. Variable nodes have features of the variable's objective coefficient, and root LP status. Constraint nodes have features of the right hand side constant $b_j$, root LP dual variables, and sense ($\le, \ge$ or $=$). There is an edge between a variable $i$ and a constraint $j$ with attribute $A_{ij}$ if variable $i$ appears in constraint $j$.
The candidate pseudo-backdoor set $\gB$ is represented as an additional binary feature for each variable node with value 1 if the variable is in $\gB$ and 0 otherwise. 
Encoding the input $(P, \gB)$ as a graph allows us to leverage state-of-the-art techniques in making predictions on graphs.


\label{sec:method:score}
We train the score model $S$ by learning to score subsets of integer variables based on their quality (i.e. runtime) as pseudo-backdoors. 
The training data contains multiple pseudo-backdoor candidates for each train MIP instance. 
%
For a MIP $P$ and two candidate subsets of integer variables $\gB_1, \gB_2$ of $P$, 
we compute the marginal ranking loss \cite{tsochantaridis2005large}  $\text{loss}(s_1, s_2, y)=\max(0, - y(s_1 - s_2) + m)$ for a given margin value $m$, where $s_1 = S(P, \gB_1 ; \theta_S), s_2 = S(P, \gB ; \theta_S)$ are the scalar score estimates and $y$ is the ranking label ($-1$ if $\gB_1$ leads to a smaller runtime than $\gB_2$,  1 othewise). Using the ranking loss trains the model to focus on distinguishing between relative performance per-instance rather than accurately modeling the absolute performance. At test time, we sample pseudo-backdoor candidate sets, score them using our model, and use the one with the best score to set branching priorities.

We learn a subsequent classifier module to determine whether to use the candidate subset or the default MIP solver. The classifier has the same architecture as the scoring model, taking as input the bipartite graph representation of the MIP $P$ and a candidate subset $\gB$, and outputting a scalar logit for binary classification. The training data contains the most promising pseudo-backdoor for each MIP instance, with a binary label indicating whether the pseudo-backdoor solved the MIP faster than the standard solver.

\section{Experiment Results}

We evaluate our method's two components on MIP instances sampled from two hardness settings of the Generalized Independent Set Problem (GISP) \cite{hochbaum1997forest}. Each hardness setting has 3 sets of 100 MIPs each, $\gD_s$ for fitting the score model, $\gD_c$ for the classification model, and $\gD_t$ for testing results. For each of the 300 MIPs, we sample 50 candidate pseudo-backdoors containing $1\%$ of the integer variables in a given MIP, with variable selection probability proportional to the variables' root LP fractionality as in \cite{dilkina2009backdoorsinitial}. 

Table \ref{tab:results} shows that the score model alone performs well on many instances. On the hard instances, scorer performs well, having $6\%$ faster runtimes than gurobi on average, and winning in 47 instances. However, scorer has $57\%$ slower runtimes on the easy instances, but still outperformed gurobi on 41 instances, demonstrating that while it has poorer performance on average, it has potential for yielding fast solve times on many instances. 
Additionally, we can see that the score model alone has much higher variability than Gurobi on the easy instances. This undesired property further motivates the inclusion of the classifier model to improve the overall performance. 
The scorer + cls model outperforms Gurobi across both MIP distributions in terms of the time distribution for MIP solving. It performs well across distributions of instances, having the lowest solve times on average and at different quantiles. The score + cls pipeline outperforms gurobi by $2\%, 8\%$ on average for easy, and hard instances respectively. 
In terms of win / tie / loss, the scorer's losses against gurobi are reduced by the classifier while its wins are retained. Furthermore, the variability on the gisp easy instances is reduced to be on par with Gurobi. 



\section{Conclusion}
In this extended abstract, we have presented a flexible method for learning and using pseudo-backdoors to improve MIP solving. 
Inspired by previous work on backdoors in MIPs, our method is comprised of two parts: an initial ranking model which scores candidate pseudo-backdoors according to how fast they will solve a given MIP, and a subsequent classification module to determine whether the selected pseudo-backdoor will indeed result in faster runtime compared with a base solver. 
Across varying difficulty levels of GISP distributions, our method finds high-quality pseudo-backdoors on unseen MIP instances. Furthermore, we find that the combination of the score model and the classification model yields consistently faster solution times compared to a state-of-the-art solver. 

\newpage
\bibliography{references}

\end{document}